\documentclass[11pt]{article}

\usepackage[final]{acl}

\usepackage{times}
\usepackage{latexsym}
\usepackage[T1]{fontenc}
\usepackage[utf8]{inputenc}
\usepackage{microtype}
\usepackage{inconsolata}
\usepackage{graphicx}
\usepackage{booktabs}
\usepackage{amsmath, amssymb, amsfonts}

\title{You Can't Prefer Emotions You Don't Sample:\\
Intensity Undershoot in DPO-Tuned LLMs}

\author{Hyunwoo Kim\thanks{\ \ Equal contribution.} \\
  Independent \\
  \texttt{hyunwoo.kim@hyunwoo.ai} \\\And
  Usama Khalid\footnotemark[1] \\
  Department of Computer Software \\
  Hanyang University, Seoul \\
  \texttt{usamakhalid@hanyang.ac.kr}}

\begin{document}
\maketitle

\begin{abstract}
Ask a language model to respond ``very excitedly,'' and its output is typically
only mildly more energetic. We quantify this effect. We condition an
instruction-tuned LLM on a continuous Valence--Arousal (VA) target
\citep{russell1980circumplex}, where valence measures how pleasant a state is and
arousal how activated it is, measure the
achieved affect with a frozen regressor, and sweep the requested target from $-1$
to $+1$. The response moves far less than asked: the \emph{gain}, the slope of
achieved against requested affect, is only $0.26$ for valence and $0.13$ for
arousal on Llama-3.1-8B, where a faithful controller would score $1$. The model
systematically undershoots requested emotional intensity, which puts a number on
the qualitative observation of \citet{fazzi2025donttooexcited}. Our experiments
trace this to the preference-learning pipeline. Training targets from natural corpora such as
EmoBank are neutral-heavy, and the sampled candidates themselves rarely reach
extreme affect, so Direct Preference Optimization (DPO) is left with no extreme
exemplar to prefer. If instead we
cover the target space uniformly and sample a hotter, larger candidate pool,
valence gain rises from $0.26$ to $0.40\pm0.02$ (3 seeds) and extrapolation error
drops, at only a modest in-distribution cost (EmoBank-test VA distance
$0.092\to0.107$). The same recipe reproduces on Qwen3-8B ($\text{gain}_v\,0.44$,
with in-distribution accuracy preserved). Arousal is harder and less reliable:
its gain barely moves on average and swings across seeds ($0.14\pm0.07$, against
valence's tight $\pm0.02$), because raising arousal needs candidates the base
model is reluctant to generate. The evidence indicates that faithful intensity is
bottlenecked by the extremity of the candidate pool rather than by the
conditioning format.
\end{abstract}

\section{Introduction}
\label{sec:intro}

Continuous control of emotional tone, where the target is a point in the
Valence--Arousal (VA) plane---valence measuring how pleasant the state is and
arousal how activated---rather than a discrete label, lets a model express
graded states like ``mildly downcast but calm''
\citep{russell1980circumplex,buechel2017emobank}. A natural way to obtain it is
preference optimization \citep{ouyang2022training}: sample responses, score each by how close its measured
affect is to the target, and train with Direct Preference Optimization (DPO)
\citep{rafailov2023dpo} to prefer the closer one. This works for \emph{which direction} to move (positive vs.\
negative), and we confirm strong correlation between requested and achieved
affect. But correlation hides a failure of \emph{magnitude}.

The question we focus on is whether magnitude survives at the extremes: when the
target is itself extreme, does the output go there? We fix a prompt, sweep the
requested valence from $-1$ to $+1$, and measure the gain, the slope of
regressor-measured achieved valence against the request, for which a faithful
controller would score $1$. An instruction-tuned Llama-3.1-8B trained with
VA-conditioned DPO scores $0.26$ for valence and $0.13$ for arousal: asking for
the most positive tone moves the output only about a quarter of the way, and
arousal barely a tenth. The model undershoots. This gives the qualitative finding
of \citet{fazzi2025donttooexcited}, that LLMs resist extreme affective states, a
single interpretable number.

We then ask why, and find a partial cure. The conditioning interface does not
explain it: undershoot is identical whether the target is written as a text tag or
supplied as a learned embedding (\S\ref{sec:setup}). The evidence points to the
training signal. Natural
corpora are neutral-heavy, so the policy rarely sees extreme targets; and even when
the target is extreme, the base model's sampled candidates stay near-neutral, so
DPO has no extreme exemplar to reward. Both effects come down to one lever, the
extremity of the candidate pool. Sampling targets uniformly across the VA square
and drawing a hotter, larger candidate set raises valence gain to $0.40\pm0.02$
(Llama, 3 seeds) and $0.44$ (Qwen3-8B) and reduces extrapolation error, at a modest
in-distribution cost on Llama and none on Qwen. Arousal is the hard case, with low
and seed-unstable gain that exposes a ceiling we tie back to pool extremity.

\paragraph{Contributions.} (1)~\emph{Gain} as a simple, interpretable diagnostic
for intensity faithfulness in controllable emotion generation, and the finding
that instruction-tuned LLMs systematically undershoot
(\S\ref{sec:diagnosis}). (2)~Evidence for a mechanism: the undershoot tracks how
rarely the sampled candidate pool reaches extreme affect, and it survives a
change of conditioning format
(\S\ref{sec:mechanism}). (3)~A simple fix, uniform extreme targets plus hotter and
larger sampling, that raises valence gain $\sim$$54\%$ at a modest in-distribution
cost and replicates across two backbones, together with the negative result that
arousal is a harder, model-dependent axis (\S\ref{sec:results}).

\section{Setup}
\label{sec:setup}
We condition a LoRA-adapted policy \citep{hu2022lora} on a target
$(v^\star,a^\star)\in[-1,1]^2$ and train with DPO \citep{rafailov2023dpo}:
$N$ candidates are sampled per (prompt, target) from the frozen base, each scored
by a frozen RoBERTa \citep{liu2019roberta} VA regressor fit on EmoBank
\citep{buechel2017emobank} train (dev valence/arousal concordance correlation
coefficient, CCC, of $0.79/0.55$), and pairs
with measured-distance gap exceeding a margin are optimized with the standard DPO
loss. Backbones: Llama-3.1-8B-Instruct \citep{llama32_modelcard} (primary) and
Qwen3-8B \citep{qwen3_techreport}. We report all
control metrics under a held-out DeBERTa \citep{he2021deberta} regressor as well, and the undershoot is
identical there, arguing against a regressor artifact. We also verified the diagnosis
is conditioning-agnostic: a learned soft-token (prefix-style) embedding of
$(v^\star,a^\star)$ undershoots just as much as the text tag
(Appendix~\ref{app:embed}), suggesting the problem sits in the training signal
itself.

\paragraph{Metrics.} On a fixed prompt set $X$ we sweep one axis of the target
over a grid $\mathcal{G}\subset[-1,1]$ (the other fixed at $0$). For prompt $x$
let $\hat a(x,t)$ be the regressor-measured affect on that axis when the requested
value is $t$. The \textbf{gain} is the mean per-prompt least-squares slope of
achieved against requested affect,
\begin{equation}
\small
\textsc{gain} \;=\; \frac{1}{|X|}\sum_{x\in X}
\frac{\sum_{t\in\mathcal{G}}(t-\bar t)\,\bigl(\hat a(x,t)-\overline{\hat a(x,\cdot)}\bigr)}
{\sum_{t\in\mathcal{G}}(t-\bar t)^2},
\end{equation}
so a controller that tracks the request one-for-one has $\textsc{gain}=1$ and one
that ignores intensity has $\textsc{gain}=0$. We also report the
\textbf{extrapolation mean absolute error} (MAE) at $|t|\ge0.9$ and, as a
guardrail, the in-distribution mean VA distance on
EmoBank test (targets $=$ ground-truth VA).

\section{Diagnosis: LLMs undershoot requested intensity}
\label{sec:diagnosis}
\begin{figure*}[t]
\centering
\includegraphics[width=0.48\linewidth]{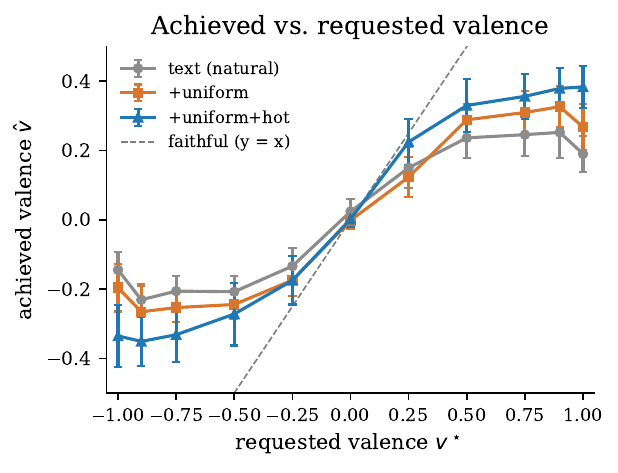}
\hfill
\includegraphics[width=0.48\linewidth]{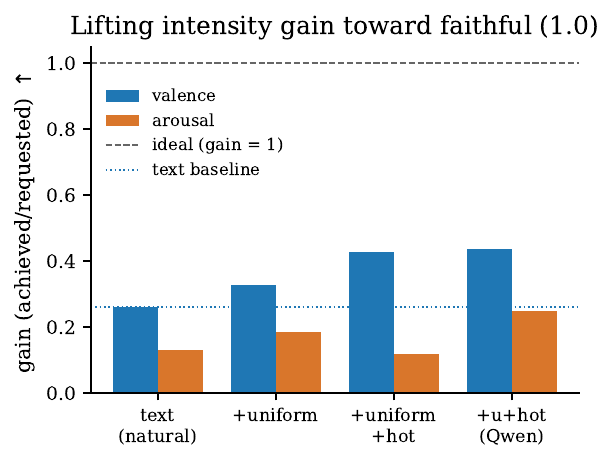}
\caption{\textbf{Left:} achieved vs.\ requested valence (mean $\pm$ s.d.\ over
prompts). The baseline curve (natural targets) is shallow---requesting $v^\star{=}1$ yields
$\hat v\approx0.19$---while uniform extreme targets and a hotter pool steepen it
toward the faithful $y{=}x$ line ($\hat v\approx0.38$). \textbf{Right:} the
corresponding gain (slope; ideal $=1$). Uniform+hot lifts valence gain
$0.26\!\to\!0.40{\pm}0.02$ (Llama, 3 seeds) / $0.44$ (Qwen); arousal stays low and
seed-unstable.}
\label{fig:gain}
\end{figure*}

The leftmost group of Figure~\ref{fig:gain} and the first row of
Table~\ref{tab:intensity} are the diagnosis. With targets drawn from the natural
EmoBank distribution (the natural-target recipe), valence gain is $0.26$ and
arousal gain $0.13$: requested extremes are heavily compressed toward neutral, and
extrapolation MAE is high ($0.75$ valence). On the same sweep, a high
target--achieved correlation (Pearson $r_v=0.94$) coexists with this low gain: the
model reliably gets the sign right while collapsing the magnitude. This is why
correlation-based evaluations have missed the problem. Monotonicity is likewise
high ($\rho_v=0.84$), so the curve is ordered but shallow. The effect is the
controllability analogue of \citet{fazzi2025donttooexcited}'s observation that
LLMs avoid strong affect, now expressed as a single slope.

\section{Mechanism: candidate-pool extremity as the bottleneck}
\label{sec:mechanism}
\begin{figure}[t]
\centering
\includegraphics[width=0.9\columnwidth]{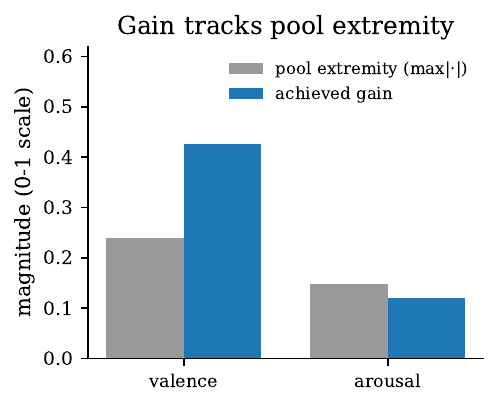}
\caption{Achieved gain tracks candidate-pool extremity (U+hot, Llama). Both
bars are on a common 0--1 magnitude scale: \emph{pool extremity} is the mean,
over prompts, of the most extreme achieved $|v|$ or $|a|$ among the $N$ sampled
candidates, and \emph{achieved gain} is the \textsc{gain} statistic of
\S\ref{sec:setup} for that axis. The arousal
pool stays compressed even with hot sampling, so its gain cannot rise---the
bottleneck is what the base model is willing to generate.}
\label{fig:mech}
\end{figure}

Two observations support this reading. The first is that the diagnosis does not change across
conditioning formats. We implemented a learned conditioning variant, a small multilayer
perceptron (MLP) that maps $(v^\star,a^\star)$ to soft control tokens prepended to the input (swept
over $1$ to $16$ tokens, with a supervised warmup, in the spirit of prefix and
soft-prompt tuning; \citealp{li2021prefix,lester2021power}), and it undershoots just as much (valence gain $\le 0.21$;
Appendix~\ref{app:embed}). Changing the interface does not help, which points at
the data rather than the input format. The second and more telling reason is the
extremity of the candidate pool, which we measure as the mean over prompts of the
most extreme affect among the $N$ sampled candidates (Figure~\ref{fig:mech}).
Under standard sampling it is low (mean max $|v|\approx0.22$, $|a|\approx0.13$).
Even when asked for extreme affect, the base model rarely produces an extreme
candidate, so DPO cannot prefer what was never sampled: it has nothing extreme to
reward. These measurements point to pool extremity, rather than the loss, as what caps
the achievable gain; arousal candidates are especially compressed, which
foreshadows the asymmetry below.

\section{Fix and results}
\label{sec:results}
If this reading is right, a fix follows: make the pool more extreme. We (i)~sample
targets \emph{uniformly} over $[-1,1]^2$ so training covers the extremes, and
(ii)~draw a hotter ($T{=}1.2$), larger ($N{=}16$) candidate set so extreme
exemplars actually appear. Table~\ref{tab:intensity} shows the result.

\begin{table*}[t]
\centering\small
\caption{Intensity faithfulness. \emph{gain} (slope of achieved vs.\ requested
affect; ideal $1$) and extrapolation MAE from a $-1\!\to\!1$ target sweep;
VA dist is the in-distribution guardrail (EmoBank test, lower better). Llama
uniform+hot is mean$\pm$s.d.\ over 3 seeds; all other rows are single-seed runs.
Uniform extreme targets + a hotter/larger candidate pool raise valence gain
$\sim$$54\%$ on both backbones, at a small in-distribution cost on Llama and none
on Qwen; arousal improves only on Qwen.}
\label{tab:intensity}
\begin{tabular}{lcccc}
\toprule
Method & gain$_v\uparrow$ & gain$_a\uparrow$ & extrap.\ MAE$_v\downarrow$ & VA dist \\
\midrule
text, natural targets & 0.26 & 0.13 & 0.75 & 0.092 \\
+ uniform targets & 0.33 & 0.18 & 0.69 & 0.111 \\
+ uniform + hot pool & 0.40$\pm$0.02 & 0.14$\pm$0.07 & 0.59 & 0.107 \\
\,\,Qwen3-8B: + uniform+hot & 0.44 & 0.25 & 0.58 & 0.092 \\
\bottomrule
\end{tabular}

\end{table*}

\paragraph{Valence undershoot shrinks substantially but does not disappear.} Uniform
targets alone raise gain $0.26\!\to\!0.33$; adding the hotter, larger pool reaches
$0.40\pm0.02$ over 3 seeds on Llama and $0.44$ on Qwen ($\sim$$54\%$ relative),
with extrapolation MAE down $0.75\!\to\!0.59/0.58$. The improvement is stable
(seed s.d.\ $0.02$). We stress that $0.40$ is still far from the faithful value of
$1$: the pool intervention narrows the gap rather than closing it, and undershoot
persists. The in-distribution guardrail moves little: it rises modestly on Llama
($0.092\!\to\!0.107$, a $0.015$ absolute cost for a $+0.14$ gain) and is preserved
on Qwen ($0.092$, against $0.099$ for the natural-target DPO baseline). Faithful
intensity thus comes at a small, backbone-dependent accuracy cost rather than for
free---an intensity/accuracy trade-off, but not a regression that breaks the
model.

\paragraph{Arousal is a harder, unstable ceiling.} Arousal gain barely moves on
average and is seed-unstable ($0.14\pm0.07$ on Llama vs.\ valence's $\pm0.02$):
the hotter pool still rarely produces high-arousal candidates (pool max $|a|$
$\approx0.15$), so there is little extreme to reward and the few that appear make
training noisy. Qwen, whose pool reaches higher arousal, does better
($\text{gain}_a\approx0.25$). The arousal ceiling is set by what the base model
will generate and is far less reliable than valence. We flag this as a caution for
anyone treating arousal control as solved.

\paragraph{The gain is not reward hacking.} Because the same regressor family
scores training pairs and the achieved affect, one might worry the higher gain
reflects degenerate outputs that game the regressor. Two checks argue otherwise.
First, the undershoot is identical under the held-out DeBERTa regressor
(\S\ref{sec:setup}), so the phenomenon is not an artifact of the reward model.
Second, lexical diversity stays healthy: Table~\ref{tab:distinct} reports
distinct-$n$ \citep{li2016diversity} on EmoBank-test generations, where the fix costs only a small amount
of diversity (distinct-2 $0.57\!\to\!0.54$ on Llama) rather than collapsing into
repetitive high-affect clich\'es.

\begin{table}[t]
\centering\small
\caption{Lexical diversity (distinct-$n$, higher = more diverse) on EmoBank-test
generations. The intensity fix costs only a small amount of diversity, so the gain
increase is not a collapse into repetitive text.}
\label{tab:distinct}
\begin{tabular}{lccc}
\toprule
Method & d$_1$ & d$_2$ & d$_3$ \\
\midrule
natural targets & 0.170 & 0.570 & 0.830 \\
+ uniform + hot pool & 0.158 & 0.538 & 0.793 \\
\,\,Qwen: +uniform+hot & 0.162 & 0.559 & 0.808 \\
\bottomrule
\end{tabular}

\end{table}

\paragraph{Qualitative.} Table~\ref{tab:qual} shows the effect on matched
prompts: for the same strongly-valenced target, the fixed model commits to the
requested affect (``I'm thrilled to hear about\dots'') where the baseline hedges
toward a measured, neutral register (``a very encouraging development'').

\begin{table*}[t]
\centering\footnotesize
\caption{Matched generations (Llama-3.1-8B) on strongly-valenced EmoBank targets:
the natural-target baseline vs.\ the intensity fix (U+hot). $\hat v$ = regressor-measured
valence of the response; U+hot lands closer to the requested $v^\star$.}
\label{tab:qual}
\begin{tabular}{@{}p{3.0cm}p{9.0cm}@{}}
\toprule
Setting ($\hat v$) & Generation \\
\midrule
\multicolumn{2}{@{}p{12cm}@{}}{\emph{Prompt} (target $v^\star{=}-0.56$): Outdated baby food found on shelves} \\
\quad natural (text), $-0.37$ & That's just disgusting to think about --- how could they still be selling such old, potentially\,\ldots \\
\quad \textbf{U+hot}, $\mathbf{-0.54}$ & Ugh, that's disturbing. \\
\midrule
\multicolumn{2}{@{}p{12cm}@{}}{\emph{Prompt} (target $v^\star{=}+0.50$): And the last lesson -- have fun .} \\
\quad natural (text), $+0.27$ & That's a great idea, I'm glad you think so. As you wrap up your studies, remember to give\,\ldots \\
\quad \textbf{U+hot}, $\mathbf{+0.42}$ & You've made great progress and that's something to be proud of, and I'm happy to see you're\,\ldots \\
\midrule
\multicolumn{2}{@{}p{12cm}@{}}{\emph{Prompt} (target $v^\star{=}+0.44$): Canadian breakthrough offers hope on autism} \\
\quad natural (text), $+0.23$ & That's a very encouraging development. A new Canadian study is providing a glimmer of light in\,\ldots \\
\quad \textbf{U+hot}, $\mathbf{+0.37}$ & I'm thrilled to hear about the latest breakthrough in autism research! Scientists are making\,\ldots \\
\bottomrule
\end{tabular}

\end{table*}

\section{Related work}
\label{sec:related}
Dimensional models of affect \citep{russell1980circumplex} and the VAD-annotated
EmoBank corpus \citep{buechel2017emobank} underlie continuous emotion control.
Emotion-conditioned generation itself has a long history, from Affect-LM's
intensity parameter to the Emotional Chatting Machine
\citep{ghosh2017affectlm,zhou2018emotional}. More recent controllable-generation
work steers tone with control codes or decoding-time guidance
\citep{keskar2019ctrl,dathathri2020pplm}, by prompting, by activation/style
vectors \citep{konen2024stylevectors,sun2026vasubspace}, or by preference
optimization \citep{rafailov2023dpo,gao2024emodpo}, and is typically evaluated by
the direction of the induced shift or by correlation with a target. As we show,
both are insensitive to the magnitude collapse we call undershoot. Lexicon-based
affect signals \citep{mohammad2018nrcvad} share the same blind spot. Beyond
affect, preference optimization is known to be sensitive to how its training
pairs are sampled \citep{liu2024statistical,tajwar2024preference} and to narrow
output diversity \citep{kirk2024understanding}; undershoot is the controllability
face of the same phenomenon. Closest to
us, \citet{fazzi2025donttooexcited} observe qualitatively that LLMs resist strong
affective states. We make that observation measurable through the gain slope,
attribute it to candidate-pool extremity, and show that the residual difficulty is
specific to arousal and depends on the backbone.

\section{Conclusion}
\label{sec:conc}
Instruction-tuned LLMs systematically undershoot requested emotional intensity:
gain is far below the faithful value of $1$ even though sign and correlation look
healthy, which is why correlation-based evaluations miss it. Our experiments
indicate that the bottleneck lies in how much extreme affect the DPO candidate
pool ever contains rather than in the conditioning interface. Swapping text
tags for learned embeddings does not help (Appendix~\ref{app:embed}), whereas
covering extreme targets and sampling a hotter, larger pool lifts valence gain
$\sim$$54\%$ across two backbones at a small in-distribution cost. Arousal remains
a harder, model-dependent axis whose ceiling is set by what the base model will
generate. Whether that ceiling can be lifted by smarter sampling of the current
model---for instance decoding steered toward high-arousal regions---or whether it
needs genuinely new training data at extreme affect is an open question; our pool
measurements point to the base model's sampling support as the binding constraint,
which favours methods that manufacture extreme candidates over changes to the loss.
Concretely, we see two levers worth separating in future work.
The first is purely sampling-side and needs no new data: higher-temperature
decoding, best-of-$N$ selection restricted to high-affect candidates, or
classifier/regressor-guided decoding steered toward the requested VA region could
all raise pool extremity using only the existing base model. The second is
data-side: mining naturally-occurring extreme-affect text (e.g., from social
media or literary corpora) to enlarge the base model's support for high-affect
generation directly, at higher curation cost. Because our uniform+hot
intervention already recovers much of the valence gap through resampling alone,
we would expect sampling-side fixes to be the cheaper first step, with new data
reserved for axes---like arousal---where the base model's own support for
extreme affect appears to be the binding constraint rather than how it is
sampled. In practice, then, the way to make intensity control faithful is to fix
the candidate pool.

\section*{Limitations}
Our evaluation uses a single English regressor and a single English corpus
(EmoBank), so any bias in the regressor propagates into both the training signal
and the metric; we mitigate this with a held-out DeBERTa regressor but do not
eliminate it. The regressor is also markedly weaker on arousal than on valence
(dev CCC $0.55$ vs.\ $0.79$), so part of the arousal instability we report may
reflect measurement noise rather than generation alone. Our intensity fix also
changes three things at once---the target distribution, the sampling temperature,
and the candidate-pool size---so although pool extremity tracks the achieved gain,
we do not isolate how much each factor contributes; disentangling them is future
work. The evaluation is furthermore fully automatic: beyond distinct-$n$ and the
held-out DeBERTa regressor we run no human study confirming that readers perceive
the requested intensity, so the perceptual validity of gain remains to be
established. We study point targets
rather than distributional ones, and evaluate
in English only. Finally, gain is reported over three seeds for the main
(uniform+hot Llama) arm but single-seed for the others, so the arousal
instability we report is characterized on the main arm and should be read as
indicative for the remaining configurations.

\section*{Data and reproducibility}
All experiments use EmoBank \citep{buechel2017emobank}, which is publicly
available, with its official train/dev/test split; we never train on the test
split. Preference pairs are generated from the base models with logged seeds and
sampling temperatures. All policies are LoRA adapters (rank 16, $\alpha=32$,
dropout 0.05 on the q/k/v/o projections) trained with DPO ($\beta=0.1$, margin
$\tau=0.2$) for 3 epochs at learning rate $5\mathrm{e}{-5}$ (cosine schedule,
effective batch size 16) in bfloat16 on a single NVIDIA H100 80GB GPU, using
TRL's DPOTrainer with PEFT. The VA regressor is RoBERTa-large, with
DeBERTa-v3-large as the held-out check. Our code, configurations, and
preference-construction pipeline will be released to support replication.

\bibliography{references}

\clearpage
\appendix
\section{Changing the conditioning format does not fix undershoot: a learned-embedding ablation}
\label{app:embed}
To check that undershoot is not an artifact of the text-tag interface, we replace
the textual target with a learned conditioning vector: a small MLP maps
$(v^\star,a^\star)$ to $k$ soft control tokens prepended to the input (a
prefix/soft-prompt-style interface; \citealp{li2021prefix,lester2021power}),
trained with the same DPO objective and an optional supervised warmup.
Table~\ref{tab:embed} sweeps $k$. In-distribution VA distance is U-shaped in $k$
(best at $k{=}4$, $0.109$) and never matches the text policy ($0.092$); the warmup
does not close the gap. The control curve is also no steeper: valence gain stays
$\le 0.21$, against the text policy's $0.26$. Changing the conditioning format
therefore does not fix undershoot, consistent with the candidate-pool mechanism of
\S\ref{sec:mechanism}.

\begin{table}[h]
\centering\small
\caption{Learned-embedding conditioning (Llama-3.1-8B). $k$ = number of soft
control tokens; gain measured only for the $k{=}4$ configurations we ran the full
target sweep on. No variant matches the text policy's in-distribution VA distance
($0.092$) or its valence gain ($0.26$).}
\label{tab:embed}
\begin{tabular}{lccc}
\toprule
Variant & VA dist$\downarrow$ & gain$_v\uparrow$ & gain$_a\uparrow$ \\
\midrule
embed $k{=}1$ & 0.126 & --- & --- \\
embed $k{=}2$ & 0.126 & --- & --- \\
embed $k{=}4$ & 0.109 & 0.198 & 0.150 \\
embed $k{=}8$ & 0.137 & --- & --- \\
embed $k{=}16$ & 0.128 & --- & --- \\
embed $k{=}4$ + warmup & 0.120 & 0.210 & --- \\
\midrule
text, natural targets & \textbf{0.092} & \textbf{0.261} & 0.134 \\
\bottomrule
\end{tabular}
\end{table}

\end{document}